%% file: plantclef2029.tex
\documentclass{llncs}
\usepackage{amssymb}
\usepackage{amsmath}
\usepackage{graphicx}
\usepackage{hyperref}
\usepackage{url}

\usepackage{float}
\floatstyle{plaintop}
\restylefloat{table}

\newcommand{\keywords}[1]{\par\addvspace\baselineskip
\noindent\keywordname\enspace\ignorespaces#1}

\usepackage{array}
\newcolumntype{C}[1]{>{\centering\let\newline\\\arraybackslash\hspace{0pt}}m{#1}}
\newcolumntype{L}[1]{>{\let\newline\\\arraybackslash\hspace{0pt}}m{#1}}

\begin{document}

\mainmatter  

\title{Overview of LifeCLEF Plant Identification task 2019: diving into data deficient tropical countries}

\titlerunning{LifeCLEF Plant Identification Task 2019}

\author{Herv\'e Go\"eau\inst{1,2}
  \and Pierre Bonnet\inst{1,2}
  \and Alexis Joly\inst{3,4}
}

\tocauthor{Herv\'e Go\"eau, Pierre Bonnet, Alexis Joly}

\institute{CIRAD, UMR AMAP, France,
\email{herve.goeau@cirad.fr, pierre.bonnet@cirad.fr}
\and 
AMAP, Univ Montpellier, CIRAD, CNRS, INRA, IRD, Montpellier, France
\and
Inria ZENITH team, France, 
\email{alexis.joly@inria.fr}
\and
LIRMM, Montpellier, France
}

\toctitle{overview lifeclef plant}

\maketitle

{\let\thefootnote\relax\footnotetext{Copyright \textcopyright\ 2019 for this paper by its authors. Use permitted under Creative Commons License Attribution 4.0 International (CC BY 4.0). CLEF 2019, 9-12 September 2019, Lugano, Switzerland.}}

\begin{abstract}
Automated identification of plants has improved considerably thanks to the recent progress in deep learning and the availability of training data. However, this profusion of data only concerns a few tens of thousands of species, while the planet has nearly 369K. The LifeCLEF 2019 Plant Identification challenge (or "PlantCLEF 2019") was designed to evaluate automated identification on the flora of data deficient regions. It is based on a dataset of 10K species mainly focused on the Guiana shield and the Northern Amazon rainforest, an area known to have one of the greatest diversity of plants and animals in the world. 
As in the previous edition, a comparison of the performance of the systems evaluated with the best tropical flora experts was carried out. This paper presents the resources and assessments of the challenge, summarizes the approaches and systems employed by the participating research groups, and provides an analysis of the main outcomes.
\end{abstract}

\keywords{LifeCLEF, PlantCLEF, plant, expert, tropical flora, Amazon rainforest, Guiana Shield leaves, species identification, fine-grained classification, evaluation, benchmark}

\section{Introduction} 
Automated identification of plants and animals has improved considerably in the last few years. In the scope of LifeCLEF 2017 \cite{joly2017lifeclef} in particular, we measured impressive identification performance achieved thanks to recent deep learning models (e.g. up to 90 \% classification accuracy over 10K species). Moreover, the previous edition in 2018 showed that automated systems are not so far from the human expertise \cite{lifeclef2018}. However, these 10K species are mostly living in Europe and North America and only represent the tip of the iceberg. The vast majority of the species in the world (about 369K species) actually lives in data deficient countries in terms of collected observations and the performance of state-of-the-art machine learning algorithms on these species is unknown and presumably much lower.\\
The LifeCLEF 2019 Plant Identification challenge (or "PlantCLEF 2019") presented in this paper was designed to evaluate automated identification on the flora of such data deficient regions. The challenge was based on a new dataset of 10K species mainly focused on the Guiana shield and the Northern Amazon rainforest, an area known to have one of the greatest diversity of plants and animals in the world. The average number of images per species in that new dataset is significantly lower than the last dataset used in the previous edition of PlantCLEF\cite{expertclef2018} (about 1 vs. 3), and many species contain very few images or may even contain only one image. To make it worse, because of the lack of illustrations of these species in the world, the data collected as a training set suffers from several properties that do not facilitate the task: many images are duplicated across different species leading to identification errors, some images do not represent plants, and many images are drawings or digitalized herbarium sheets that may be visually far from field plant images. The test set, on the other hand, does not present this type of noisy and biased content since it is composed only of expert data identified in the field with certainty. As these data have never been published before, there is also no risk that they belong to the training set.\\
As in the 2018-th edition of PlantCLEF, a comparison of the performance of the systems evaluated with the best tropical flora experts was carried out for PlantCLEF 2019. In total, 26 deep-learning systems implemented by 6 different research teams were evaluated with regard to the annotations of 5 experts of the targeted tropical flora. This paper presents more precisely the resources and assessments of the challenge, summarizes the approaches and systems employed by the participating research groups, and provides an analysis of the main outcomes.

\section{Dataset}

\subsection{Training set}
\label{training}
\begin{figure}
\centering
\includegraphics[width=0.5\linewidth]{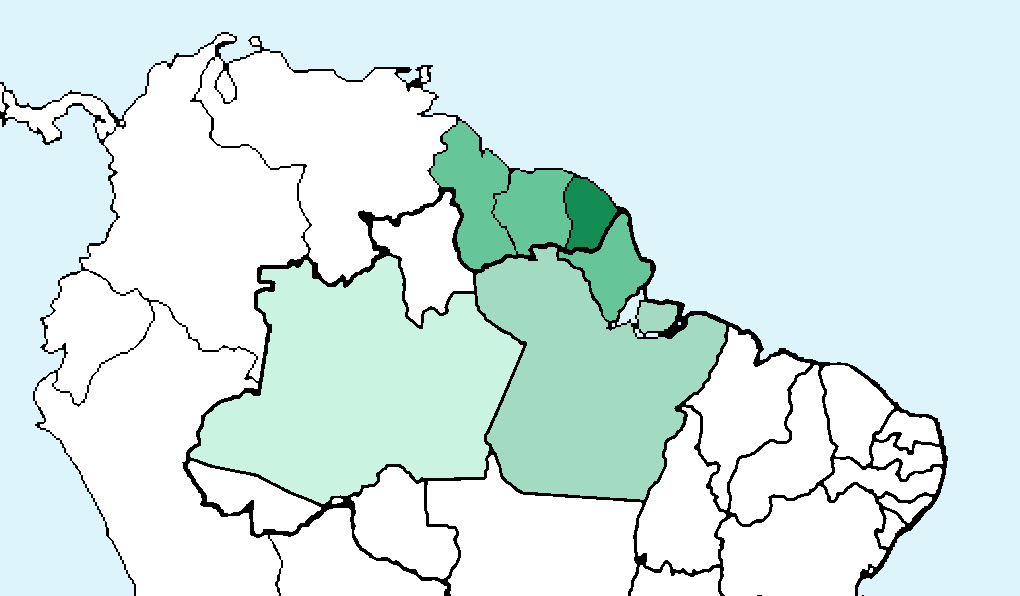}
\caption{Regions of origin of the 10k species selected for PlantCLEF 2019: French Guiana, Suriname, Guyana, Brazil (states of Amapa, Para, Amazonas)}
\label{fig:plantclef2019speciesmap}
\end{figure}

We provided a new training data set of 10K species mainly focused on the Guiana shield and the Amazon rainforest, known to be one of the largest collection of living plants and animal species in the world (see figure\ref{fig:plantclef2019speciesmap}). As for the two previous years, this training data was mainly aggregated by bringing together images from complementary types of available sources, including expert data from the international platform Encyclopedia of Life (EoL\footnote{\url{https://eol.org/}}) and images automatically retrieved from the web using industrial search engines (Bing and Google) that were queried with the binomial Latin name of the targeted species. Details numbers of images and species per sub-dataset are provided in Table \ref{table:SDD_dataset}. A large part of the images collected come from trusted websites, but they also contain a high level of noise. It has been shown in previous editions of LifeCLEF, however, that training deep learning models on such raw big data can be as effective as training models on cleaner but smaller expert data \cite{plantclef2017}, \cite{expertlifeclef2018}. The main objective of this new study was to evaluate whether this inexpensive methodology is applicable to the case of tropical floras that are much less observed and therefore much less present on the web.

One of the consequences of this change in target flora is that the average number of images per species is much lower (about 1 vs. 3). Many species contain only a few images and some of them even contain only 1 image. On the other hand, few common species are still associated with several hundreds of pictures mainly coming from EOL. In addition to this scarcity of data, the use of web search engines to collect the images generates several types of noise:\\
\\
\textbf{Duplicate images and taxonomic noise:} web search engines often return the same image several times for different species. This typically happens when an image is displayed in a web page that contains a list of several species. For instance, in the Wikipedia web page of a genus, all child species are usually listed but only a few of them are illustrated with a picture. As a consequence, the available pictures are often retrieved for several species of the genus and not only the correct one. We call this phenomenon \textit{taxonomic noise}. The resulting label errors are problematic but they can paradoxically have a certain usefulness during the training. Indeed, species of the same genus often have a number of common morphological traits, which can be visually similar. For the less illustrated species, it is therefore often more cost-effective to keep images of closely related species rather than not having images at all. Technically, to help managing this taxonomic noise, the duplicate images were replicated in the directories of each species they belong to, but the same image name was used everywhere.\\
\\
\textbf{Non-photographic images of the plant (herbarium, drawings):} because of data scarcity, it often occurs that the only images available on the web for a given species are not photographs but rather digitized herbarium sheets or drawings from academic books. This typically happens for very rare or poorly observed species. For the most extreme cases, these old testimonies, sometimes more than a century old, are the only existing data. The usefulness of these data for learning is again ambivalent. They can be visually very different from a photograph of the plant but they still contain a rich information about the appearance of the species.\\
\\
\textbf{Atypical photographs of the plant:} a number of images are related to the target species but do not directly represent it. Typically, it can be a landscape related to the habitat of the species, a photograph of the dissection of plant organs, a handful of seeds on a blank sheet of paper, or a microscopic view.\\
\\
\textbf{Non-plant images:} search engines sometimes return images that are not plants and that have only a very indirect link with the target species: medicines, animals, mushrooms, botanists, handcrafted-objects, logos, maps, ethnic (food, craftsmanship), etc.

\begin{table}[t!]\footnotesize
    \centering
	\caption{Description of the PlantCLEF training sub-datasets.}
	\resizebox{\columnwidth}{!}{
	\begin{tabular}{ c  c c c  c  c }
		\hline
		Sub-dataset names  & \# Images &  \# species  & Data source\\  \hline
		PlantCLEF 2019 EOL & 58,619 & 4,197 & Encyclopedia of Life\\
		PlantCLEF 2019 Google & 68,225 & 6,277 & Automatically retrieved by Google web search engine\\
		PlantCLEF 2019 Bing & 307,407 & 8,681 & Automatically retrieved by Bing web search engine\\ \hline
		Total & 434,251 & 10,000 & \\\hline
	\end{tabular} \vspace{-1.5em}
	\label{table:SDD_dataset}
	}
\end{table}


\subsection{Test set}
Unlike the training set, the labels of the images in the test set are of very high quality to ensure the reliability of the evaluation. It is composed of 742 plant observations, all of which have been identified in the field by one of the best experts on the flora in question. Marie-Françoise Pr\'evost \cite{delprete2013marie}, the author of this test set, is a french botanist, who has spent more than 40 years to study  French Guiana flora. She has an extensive field work experience, and she has contributed a lot to improve our current knowledge of this flora, by collecting numerous herbarium specimens of great quality. Ten plant epithets have been dedicated to her, which illustrates the acknowledgement of the taxonomists community to her contribution to the tropical botany. For the re-annotation experiment by the other 5 human experts, only a sub-set of 117 of these observations was used.



\section{Task Description}
The goal of the task was to identify the correct species of the 742 plants of the test set. For every plant, the evaluated systems had to return a list of species, ranked without ex-aequo. Each participating group was allowed to submit up to 10 \textit{run files} built from different methods or systems (a \textit{run file} is a formatted text file containing the species predictions for all test items).\\

The goal of the task was exactly the same for the 5 human experts, except that we restricted the number of their responses to 3 species per test item to reduce their effort. The list of possible species was provided to them.\\

The main evaluation metric for both the systems and the humans was the \textit{Top1 accuracy}, \textit{i.e.} the percentage of test items for which the right species is predicted in first position. As complementary metrics, we also measured the \textit{Top3} accuracy, \textit{Top5} accuracy and the Mean Reciprocal Rank (MRR), defined as the mean of the multiplicative inverse of the rank of the correct answer:
$$ \text{MRR} : \frac{1}{Q} \sum_{q=1}^Q \frac{1}{\text{rank}_q} $$ 

\section{Participants and methods}
\label{particp}
167 participants registered for the PlantCLEF challenge 2019 and downloaded the data set, but only 6 research groups succeeded in submitting \textit{run files}. Details of the methods are developed in the individual working notes of most of the participants (Holmes \cite{NeuonAI2019}, CMP \cite{CMP2019},  MRIM-LIG \cite{MRIMLIG2019}). We provide hereafter a synthesis of the runs of the best performing teams:\\ 
\\
\textbf{CMP, Dept. of Cybernetics, Czech Technical University in Prague, Czech Republic, 7 runs, \cite{CMP2019}}: this team used an ensemble of 5 Convolutional Neural Networks (CNNs) based on 2 state-of-the-art architectures (Inception-ResNet-v2 and Inception-v4). The CNNs were initialized with weights pre-trained on the dataset used during ExpertCLEF2018 \cite{CMP2018expert} and then fine-tuned with different hyper-parameters and with the use of data augmentation (random horizontal flip, color distortions and random crops). Further performance improvements were achieved by adjusting the CNN predictions according to the estimated change of the classes distribution between the training set and the test set. The running averages of the learned weights were used as the final model and the test was also processed with data augmentation (3 central crops at various scales and their mirrored version). 
Regarding the data used for the training, the team decided to remove all images estimated to be non ﬂoral data based on the classiﬁcation of a dedicated VGG net. This had the effect of removing 300 species and about 5,500 pictures. An important point is that additional training images were downloaded from the GBIF platform\footnote{\url{https://www.gbif.org/}}, in order to fill the missing species and enrich the other species. This increased considerably the training dataset with 238,009 new pictures of good quality totalling then 666,711 pictures. Unfortunately, the participants did not submit a run without the use of this new training data, so that it not possible to measure accurately the impact of this addition. The participant's focus was rather on evaluating different prior distribution of the classes (Uniform, Maximum Likelihood Estimate, Maximum a Posteriori) to modify predictions in order to soften the impact of the highly unbalanced distribution of the training set.\\
\\
\textbf{Holmes, Neuon AI, Malaysia, 3 runs, \cite{NeuonAI2019}}: 
This team used the same CNN architectures than the CMP team (Inception-v4 and Inception-ResNet-v2). In their case, however, the CNNs were initialized with weights pre-trained on ImageNet rather than ExpertCLEF2018 \cite{CMP2018expert} and they did not use any additional training data. An original feature of their system was the introduction of a multi-task classification layer, allowing to classify the images at the genus and family levels in addition to the species. Complementary, this team spent some efforts to clean the dataset. The 154,627 duplicate pictures were removed and they automatically removed 15,196 additional near-duplicates based on a cosine similarity in the feature space of the last layer of Inception-V4. Finally, they removed 13,341 non plant images automatically detected by using a plant vs. non plant binary classifier (also based on Inception-V4). Overall, the whole training set was decreased by nearly 42\% of the images (whereas the CMP team increased it by nearly 53\%).\\
Cross-validation experiments conducted by the authors \cite{NeuonAI2019} show that removing the duplicates and near-duplicates may allow to gain 4 points of accuracy. In contrast, removing the non plant pictures does not provide any improvement. The introduction of the multi-task classifier at the different taxonomic levels is shown to provide one to two more points of accuracy.\\
\\
\textbf{MRIM, LIG, France, 10 runs, \cite{MRIMLIG2019}}: this team based all the runs on DenseNet, another state-of-art CNN architecture which has the advantage to have a relatively low number of parameters compared to other popular CNNs. They increased the initial model with a non-local block with the idea to model interpixels correlations from diﬀerent positions in the feature maps. They used a set of data augmentation processes including random resize, random crop, random flip and random brightness and contrast changes. To compensate the class imbalance, they made use of oversampling and under-sampling strategies. Their cross-validation experiments did show some significant improvements but these benefits were not confirmed on the final test set, probably because of the cross-validation methodology (based on a subset of only 500 species among the most populated ones).\\
\\
The three other remaining teams did not provide an extended description of their system. According to the short description they provided, the datvo06 team from Vietnam (1 run) used a similar approach to the MRIM team (DenseNet), the Leowin team from India (2 runs) used Random Forest Boosted on the features extracted from a ResNet, and the MLRG SSN team from India (3 runs), used a ResNet 50 trained for 100 epochs with stratification of batches.

\section{Results}

\begin{figure}
\centering
\includegraphics[width=0.9\linewidth]{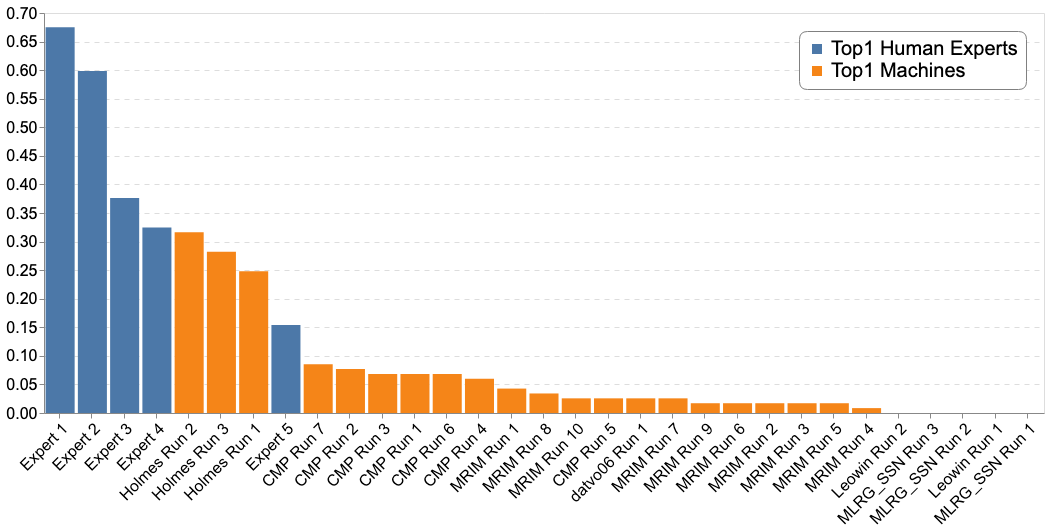}
\vspace{-5mm}
\caption{Scores between Experts and Machine}
\label{fig:PlantCLEF2019ScoresMvsM}
\end{figure}

\begin{figure}
\centering
\includegraphics[width=\linewidth]{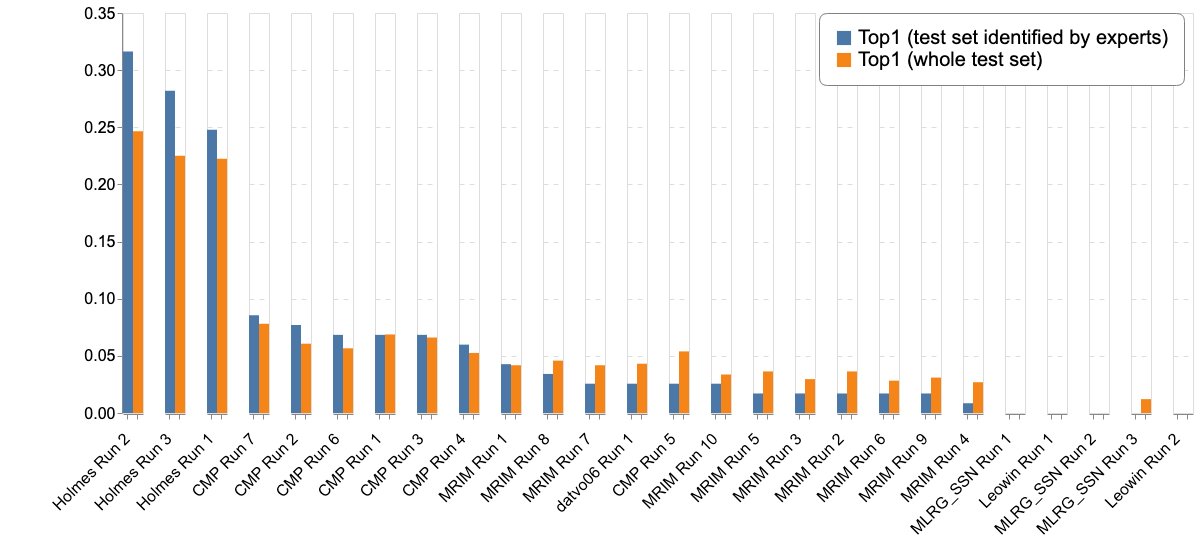}
\vspace{-5mm}
\caption{Scores achieved by all systems evaluated on the two test sets}
\label{fig:PlantCLEF2019OfficialScore}
\end{figure}

The detailed results of the evaluation are reported in Table \ref{tab:rawresults}. Figure \ref{fig:PlantCLEF2019ScoresMvsM} gives a more graphical view of the comparison between the human experts and the evaluated systems (on the dedicated test subset). Figure \ref{fig:PlantCLEF2019OfficialScore}, on the other hand, provides a comparison of the evaluated systems on the whole test set.
\input{tables/results_table.tex}

The main outcomes we can derive from that results are the following ones:\\
\\
\\
\\
\textbf{A very difficult task, even for experts:} none of the botanist correctly identified all observations. The top-1 accuracy of the experts is in the range $0.154-0.675$ with a median value of $0.376$. It illustrates the difficulty of the task, especially when reminding that the experts were authorized to use any external resource to complete the task, with Flora books in particular. It shows that a large part of the observations in the test may not contain enough information to be identified with high confidence. The complete identification may actually rely on other information such as the root shape, the smell of the plant, type of habitat, the feeling of touch from certain parts, or the presence of organs or feature that were not photographed of visible on the pictures. Only two experts with an exceptional field expertise were able to correctly identify more than $60\%$ of the observations. The other ones correctly identified less than $40\%$.\\
\\
\textbf{Tropical flora is much more difficult to identify}. Results are significantly lower than the previous edition of LifeCLEF confirming the assumption that the tropical flora is inherently more difficult to identify than the more generalist flora. The best accuracy obtained by an expert is 0.675 for the tropical flora whereas it was 0.96 for the flora of temperate regions considered in 2018\cite{expertclef2018}. Comparison of medians (0.376 vs 0.8) and minimums (0.154 vs 0.613) over the two years further highlights the gap. This can be explained by the fact that (i) there is in general much more diversity in tropical regions compare to temperate ones, for a same reference surface, (ii) tropical plants in high rainforests, are much less accessible to humans who have much more difficulties to improve their knowledge on these ecosystems, (iii) the volume of available resources (including herbarium specimens, books, web sites) is much less important on that floras.\\
\\
\textbf{Deep learning algorithms were defeated by far by the best experts}. The best automated system is half as good as the best expert with a gap of 0.365, whereas last year the gap was only 0.12. Moreover, there is a strong disparity in results between participants despite the use of popular and recent CNNs (DensetNet, ResNet, Inception-ResNet-V2, Inception-V4), while during the last four PlantCLEF editions the homogenization of high results forming a "skyline" had often been observed. These differences in accuracy can be explained in part by the way participants managed the training set. Although previous work had shown the effectiveness of training from noisy data \cite{krause2016unreasonable},\cite{plantclef2017}, most teams considered that the training dataset was too noisy and too imbalanced. They made consistent efforts for removing duplicates pictures (Holmes), for removing non plant pictures (Holmes, CMP), for adding new pictures (CMP), or for reducing the classes imbalance with smoothed re-sampling and other data sampling schemes (MRIM). None of them attempted to simply run one of their models on the raw data (as usually done in previous years). So that it is is not possible to conclude on the benefit of such filtering methods this year.

\section{Complementary results}
\begin{figure}
\centering
\includegraphics[width=\linewidth]{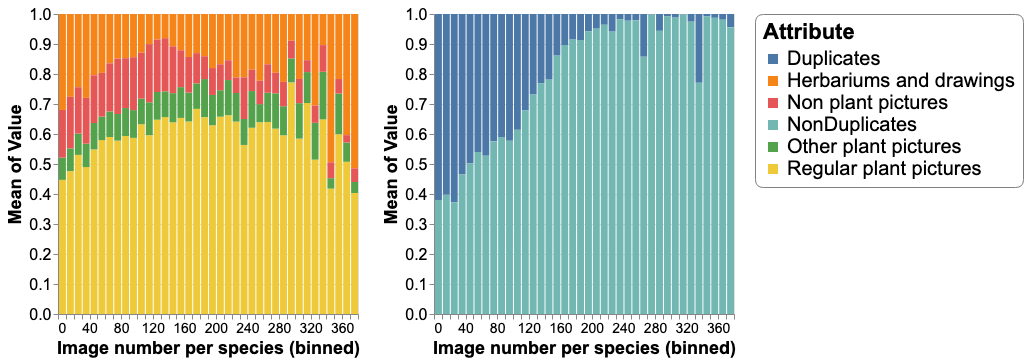}
\vspace{-5mm}
\caption{Average proportion of each noise as a function of the number of training images per species (binned with a step size of 10 pictures)}
\label{fig:plantclef2019datasetestimationdistrib}
\end{figure}

\textbf{Extending the training set with herbarium data may provide significant improvements.} As mentioned in section \ref{particp}, the CMP team considerably extended the training set by adding more than 238k images of the GBIF platform, the vast majority of these images coming from the digitization of herbarium collections. The performance of their system during the official evaluation was not that good, but unfortunately, this was mainly due to a bug in the formatting of their submissions. The corrected version of their submissions were evaluated after the end of the official challenge and did achieve a top-1 accuracy of 41\%, 10 points more than the best model of the evaluation and 3 points more than the third human expert. It is likely that this high performance gain is mostly due to the use of the additional training data. This opens up very interesting perspectives for the massive use of herbarium collections, which are being digitized at a very fast pace worldwide.\\
\\
\textbf{Estimation of class purity in the training dataset.}
In order to evaluate how the different types of noise (\textit{herbariums \& drawings}, \textit{other plant pictures} and \textit{non plant pictures}) affect the training set, we computed some statistics based on a semi-automated classification of the training set. More precisely, we repeatedly annotated some images by hand, trained dedicated classifiers and predicted the missing labels. Figure \ref{fig:plantclef2019datasetestimationdistrib} (left side) displays the average proportion of each noise as a function of the number of training images per species. Complementary, on the right side, we display the average proportion of duplicates still as a function of the number of training images per species. If we look at the \textit{herbariums \& drawings}, it can be seen that their proportion significantly decreases up to 150 images per species and then increases again for the most populated species. This evolution has to be correlated with the average proportion of web vs EoL data in the training set. Indeed, the number of web images per species was limited to 150 so that the species with large amounts of training data are mainly illustrated by EoL images. This data is highly trusted in terms of species labels but the figure shows that it contains a high proportion of herbarium sheets. Concerning the two other types of noise, the proportion of \textit{other plant pictures} is globally increasing likely because this type of picture is also more represented in the EoL data. In contrast, the proportion of \textit{non plant pictures} is decreasing above 150 images per species which means that this kind of noise is lower in EoL. Concerning the duplicates, it can be seen that their proportion is strongly decreasing with the number of images per species. This means that the absolute number of duplicates is quite stable over all species but that its relative impact is much more important for species having scarce training data.\\

\begin{figure}
\centering
\includegraphics[width=\linewidth, height=0.9\textheight]{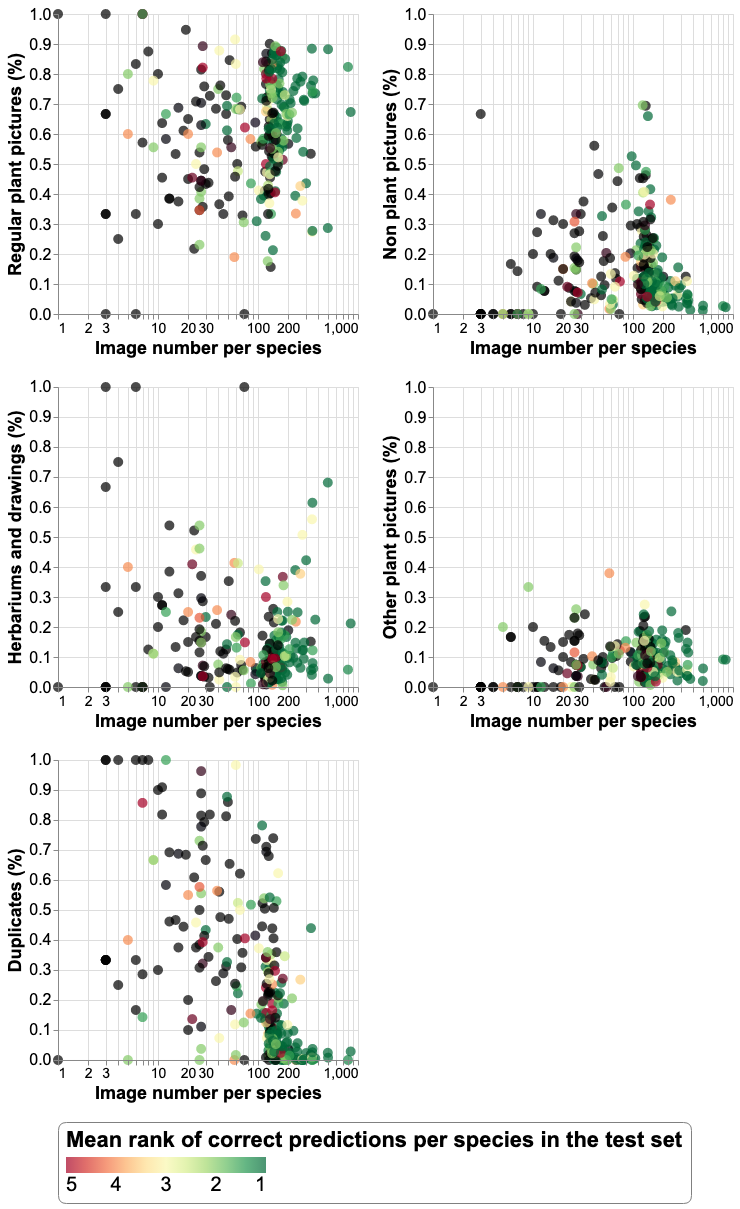}
\caption{Mean rank for each species as a function of the number of training images for this species and the estimated proportions of the different noises. The mean ranks greater than or equal to 6 are shown in black. }
\label{fig:plantclef2019datasetestimationwithtest}
\end{figure}

\textbf{Average accuracy of the best evaluated systems as a function of the number of training images per species}
To analyze the impact of the different types of noise on the prediction performance, we computed the species-wise performance of a fusion of the best run of each team (focusing on the three teams who obtained the best results CMP, Holmes and MRIM). This was done by first averaging the scores returned by each system and then by computing the mean rank of the correct answer over all the test images of a given species. Figure \ref{fig:plantclef2019datasetestimationwithtest} displays this mean rank for each species (using a color code, see legend) as a function of the number of training images for this species and the estimated proportions of the different noises. The following conclusions can be derive from these graphs:
\begin{enumerate}
    \item the more images, the better the performance: without surprise, all graphs show that the mean rank of the correct species improves with the number of images.
    \item the presence of \textit{non plant pictures} only affects species with few training data (as shown in the second sub-graph). Well populated species seem to be well recognized even with a very high proportion of \textit{non plant pictures}.
    \item the presence of \textit{herbarium data} and \textit{other plant pictures} is not conclusive: as discussed in section \ref{training}, these two types of contents are ambivalent. They may bring some useful information but they may also disrupt the model. The graphs of \ref{fig:plantclef2019datasetestimationwithtest} do not allow to conclude on this point. 
    \item a too high proportion of duplicates (above 20\%) significantly degrades the results, even for species having between 30 and 200 images. 
\end{enumerate}
\section{Conclusion}
This paper presented the overview and the results of the LifeCLEF 2019 plant identification challenge following the eight previous editions conducted within CLEF evaluation forum. The results reveal that the identification performance on Amazonian plants is considerably lower than the one obtained on temperate plants of Europe and North America. The performance of convolutional neural networks fall due to the very low number of training images for most species and the higher degree of noise that is occurring in such data. Human experts themselves have much more difficulty identifying the tropical specimens evaluated this year compared to the more common species considered in previous years. This shows that the small amount of data available for these species is correlated with the lowest overall knowledge we have of them. An interesting perspective for the future is to consider herbarium data as one solution to overcome the lack of data. This material, collected by botanists for centuries, is the most comprehensive knowledge that we have to date for a very large number of species on earth. Thus, their massive ongoing digitization represents a great opportunity. Nevertheless, they are very different from field photographs, their use will thus pose challenging domain adaptation problems.\\
\\
\\
\\
\\
\\
\textbf{Acknowledgements}
We would like to thank very warmly Julien Engel, R\'emi Girault, Jean-Fran\c{c}ois Molino and the two other expert botanists who agreed to participate in the task on plant identification. We also we would like to thank the University of Montpellier and the Floris'Tic project (ANRU) who contributed to the funding of the 2019-th edition of LifeCLEF.

\bibliographystyle{splncs03}

\end{document}

%% file: tables/results_table.tex
\begin{table}
    \centering
    \vspace{3mm}
    \begin{tabular}{|>{\bfseries}C{31mm}|>{\bfseries}C{11.5mm}|>{\bfseries}C{11mm}|C{10mm}|C{10mm}|C{10mm}|C{10mm}|C{10mm}|}
    \hline
    Team run	&	Top1 Expert	&	Top1 Whole	&	Top3 Expert	&	Top5 Expert	&	Top5 Whole	&	MRR Expert	&	MRR Whole	\\
    \hline
    \hline
Holmes Run 2	&	0,316	&	0,247	&	0,376	&	0,419	&	0,357	&	0,362	&	0,298	\\
Holmes Run 3	&	0,282	&	0,225	&	0,359	&	0,376	&	0,321	&	0,329	&	0,274	\\
Holmes Run 1	&	0,248	&	0,222	&	0,325	&	0,368	&	0,325	&	0,302	&	0,269	\\
CMP Run 7	&	0,085	&	0,078	&	0,145	&	0,197	&	0,168	&	0,124	&	0,111	\\
CMP Run 2	&	0,077	&	0,061	&	0,145	&	0,188	&	0,162	&	0,117	&	0,097	\\
CMP Run 6	&	0,068	&	0,057	&	0,154	&	0,188	&	0,163	&	0,112	&	0,096	\\
CMP Run 1	&	0,068	&	0,069	&	0,145	&	0,171	&	0,158	&	0,107	&	0,099	\\
CMP Run 3	&	0,068	&	0,066	&	0,128	&	0,188	&	0,156	&	0,110	&	0,099	\\
CMP Run 4	&	0,060	&	0,053	&	0,128	&	0,162	&	0,160	&	0,097	&	0,090	\\
MRIM Run 1	&	0,043	&	0,042	&	0,051	&	0,060	&	0,088	&	0,055	&	0,063	\\
MRIM Run 8	&	0,034	&	0,046	&	0,068	&	0,103	&	0,102	&	0,057	&	0,068	\\
MRIM Run 7	&	0,026	&	0,042	&	0,085	&	0,094	&	0,096	&	0,053	&	0,065	\\
datvo06 Run 1	&	0,026	&	0,043	&	0,051	&	0,060	&	0,086	&	0,041	&	0,061	\\
CMP Run 5	&	0,026	&	0,054	&	0,085	&	0,085	&	0,119	&	0,050	&	0,078	\\
MRIM Run 10	&	0,026	&	0,034	&	0,068	&	0,068	&	0,085	&	0,047	&	0,057	\\
MRIM Run 5	&	0,017	&	0,036	&	0,043	&	0,077	&	0,082	&	0,039	&	0,058	\\
MRIM Run 3	&	0,017	&	0,030	&	0,060	&	0,077	&	0,088	&	0,043	&	0,054	\\
MRIM Run 2	&	0,017	&	0,036	&	0,043	&	0,077	&	0,082	&	0,039	&	0,058	\\
MRIM Run 6	&	0,017	&	0,028	&	0,051	&	0,077	&	0,078	&	0,037	&	0,049	\\
MRIM Run 9	&	0,017	&	0,031	&	0,043	&	0,068	&	0,088	&	0,039	&	0,055	\\
MRIM Run 4	&	0,009	&	0,027	&	0,060	&	0,077	&	0,077	&	0,038	&	0,049	\\
MLRG SSN Run 1	&	0,000	&	0,000	&	0,000	&	0,000	&	0,000	&	0,000	&	0,000	\\
Leowin Run 1	&	0,000	&	0,000	&	0,000	&	0,000	&	0,001	&	0,000	&	0,000	\\
MLRG SSN Run 2	&	0,000	&	0,000	&	0,000	&	0,000	&	0,000	&	0,000	&	0,000	\\
MLRG SSN Run 3	&	0,000	&	0,012	&	0,000	&	0,009	&	0,027	&	0,004	&	0,021	\\
Leowin Run 2	&	0,000	&	0,000	&	0,000	&	0,000	&	0,001	&	0,000	&	0,000	\\
\hline
Expert 1	&	0,675	&	-	&	0,684 	&	0,684	&	-	&	0,679	&	-	\\
Expert 2	&	0,598	&	-	&	0,607	&	0,607	&	-	&	0,603	&	-	\\
Expert 3	&	0,376	&	-	&	0,402	&	0,402	&	-	&	0,389	&	-	\\
Expert 4	&	0,325	&	-	&	0,530	&	0,530	&	-	&	0,425	&	-	\\
Expert 5	&	0,154	&	-	&	0,154	&	0,154	&	-	&	0,154	&	-	\\
\hline
\end{tabular}
\caption{Results of the LifeCLEF 2019 Plant Identification Task}
\label{tab:rawresults}
\end{table}